\documentclass[11pt,a4paper]{article}
\usepackage{times,latexsym}
\usepackage{url}
\usepackage[T1]{fontenc}

\usepackage[acceptedWithA]{tacl2021v1}

\usepackage{xspace,mfirstuc,tabulary}

\newif\iftaclinstructions
\taclinstructionsfalse 
\iftaclinstructions
\renewcommand{\confidential}{}
\renewcommand{\anonsubtext}{(No author info supplied here, for consistency with
TACL-submission anonymization requirements)}
\newcommand{\instr}
\fi

\iftaclpubformat 

\else

\fi

\usepackage{amsmath}
\usepackage{amssymb}
\usepackage{float}
\usepackage{framed}
\usepackage[makeroom]{cancel}
\usepackage{graphicx}
\usepackage{booktabs}
\usepackage{wasysym} 
\usepackage{xcolor}
\usepackage{xspace}
\usepackage{array}
\usepackage{multirow}

\definecolor{mygreen}{RGB}{0, 102, 0}
\newcommand{\added}[1]{#1}

\newcommand{\leftquestion}{\textcolor{red}{\textbf{Q}}\xspace}
\newcommand{\rightquestion}{\textcolor{mygreen}{\textbf{Q'}}\xspace}

\newcommand{\mpt}{\phantom{.}---\phantom{5}} 
\newcommand{\p}{\phantom{5}} 
\newcommand{\x}{\phantom{.}} 

\newcommand{\data}{MuSiQue\xspace} 
\newcommand{\dataans}{MuSiQue-Ans\xspace}
\newcommand{\datafull}{MuSiQue-Full\xspace}
\newcommand{\condition}{MuSiQue\xspace}
\newcommand{\hpqa}{HotpotQA\xspace}
\newcommand{\twowikifull}{2WikiMultihopQA\xspace}

\newcommand{\iconl}{{\LARGE \textcolor{orange}{\twonotes}\,}}
\newcommand{\iconm}{{\large \textcolor{orange}{\twonotes}\,}}
\newcommand{\iconpm}{\phantom{{\large \textcolor{orange}{\twonotes}\,}}}

\newcommand{\hq}{HQ\xspace}
\newcommand{\tw}{2W\xspace}
\newcommand{\ma}{\iconm-Ans\xspace}
\newcommand{\mf}{\iconm-Full\xspace}

\newcommand{\bridgeone}[1]{\textcolor{brown}{\textbf{#1}}}
\newcommand{\bridgetwo}[1]{\textcolor{magenta}{\textbf{#1}}}
\newcommand{\bridgethree}[1]{\textcolor{blue}{\textbf{#1}}}
\newcommand{\bridgefour}[1]{\textcolor{teal}{\textbf{#1}}}

\newcommand{\singlemaskedqn}[2]{$q_{#1}^{m_{#2}}$}

\newcommand{\msinglemaskedqn}[2]{q_{#1}^{m_{#2}}}

\newcommand{\munmaskedqn}[1]{q_{#1}}

\newcommand{\mh}{multihop\xspace}
\newcommand{\sh}{single-hop\xspace}

\newcommand{\SH}{Single-hop\xspace}
\newcommand{\Mh}{Multihop\xspace}

\newcommand{\overeat}[1]{}
\newcommand{\taclvone}[1]{}

\title{
    \iconl
    MuSiQue: Multihop Questions via Single-hop Question Composition
}

\author{
  Harsh Trivedi$^\diamond$\ \ \ \
  Niranjan Balasubramanian$^\diamond$\ \ \ \
  Tushar Khot$^\dagger$\ \ \ \
  Ashish Sabharwal$^\dagger$\\
  \\
  $^\diamond$Stony Brook University, Stony Brook, U.S.A.\\
  \texttt{\small \{hjtrivedi,niranjan\}@cs.stonybrook.edu}\\
  \ \\
  $^\dagger$Allen Institute for AI, Seattle, U.S.A.\\
  \texttt{\small \{tushark,ashishs\}@allenai.org}
}

\date{}

\begin{document}
\maketitle


\begin{abstract}
Multihop reasoning remains an elusive goal as existing multihop benchmarks are known to be largely solvable via shortcuts. Can we create a question answering (QA) dataset that, by construction, \emph{requires} proper multihop reasoning? To this end, we introduce a bottom-up approach that systematically selects composable pairs of single-hop questions that are connected, i.e., where one reasoning step critically relies on information from another. This bottom-up methodology lets us explore a vast space of questions and add stringent filters as well as other mechanisms targeting connected reasoning. It provides fine-grained control over the construction process and the properties of the resulting $k$-hop questions. We use this methodology to create MuSiQue-Ans, a new multihop QA dataset with 25K 2-4 hop questions. Relative to existing datasets, MuSiQue-Ans is more difficult overall (3x increase in human-machine gap), and harder to cheat via disconnected reasoning (e.g., a single-hop model has a 30 point drop in F1). We further add unanswerable contrast questions to produce a more stringent dataset, MuSiQue-Full. We hope our datasets will help the NLP community develop models that perform genuine multihop reasoning.\footnote{Code and datasets available at \url{https://github.com/stonybrooknlp/musique}.}
\end{abstract}

\section{Introduction}

\begin{figure}[t]
    \centering
	\includegraphics[width=0.47\textwidth]{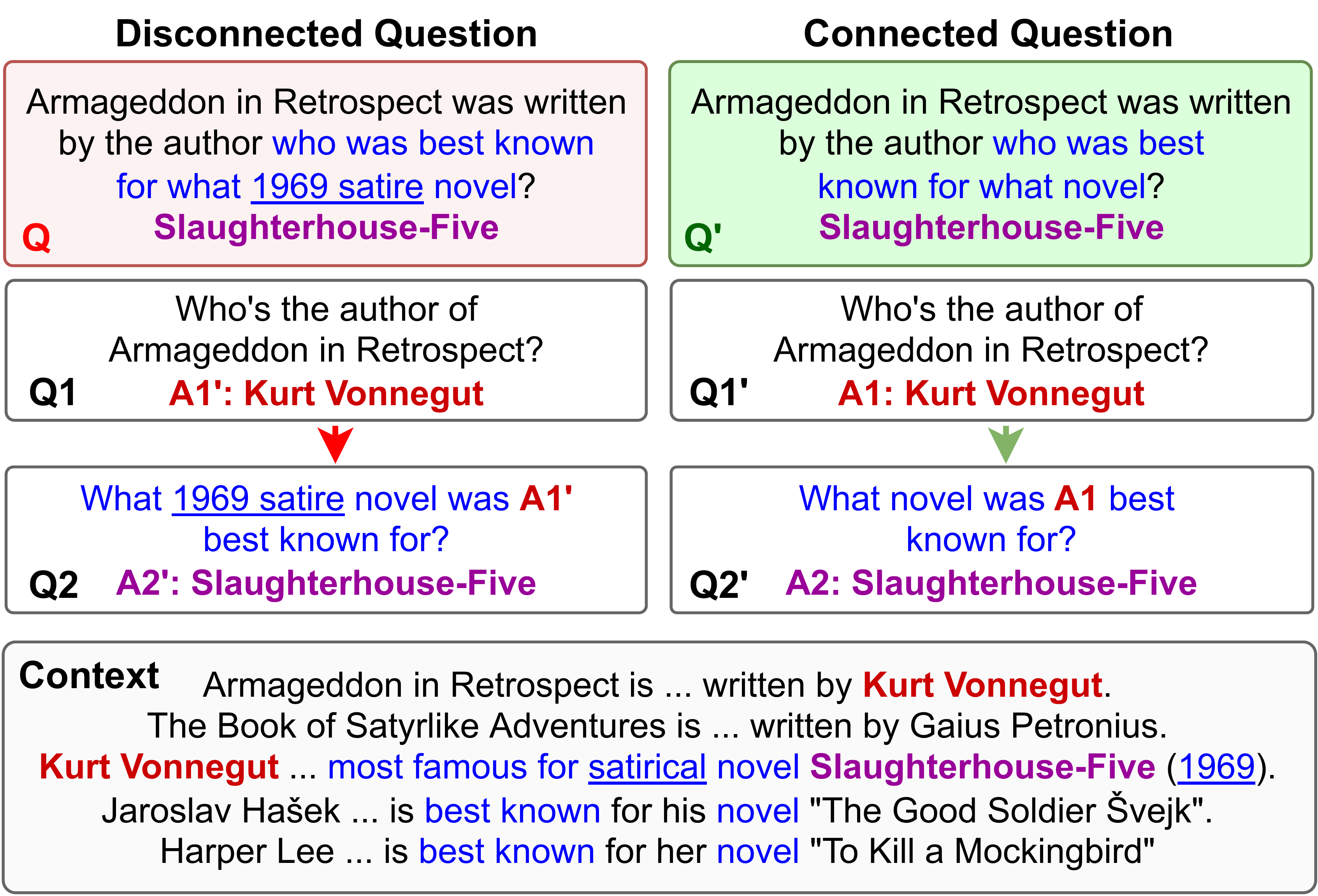}
	\caption{
    Generating connected \mh questions by composing carefully chosen pairs of \sh questions. \textbf{Left}: A HotpotQA question that would have been filtered out by our approach for not requiring connected reasoning; it can be answered using just Q2 \emph{without} knowing the answer to Q1 (since there is only one person mentioned in the context as being best known for a satirical novel). \textbf{Right}: A connected question that forces models to reason through both intended hops (since there are multiple people mentioned in the context as being best known for some novel).
    }
	\label{fig:introduction}
\end{figure}

\Mh QA datasets are designed to support the development and evaluation of models that perform multiple steps of reasoning
in order to answer a question. Recent work, however, shows that on existing datasets, models often need not even {\em connect} information across all supporting facts,\footnote{For example, they often don't even use information from one supporting fact to select another.} because they can exploit reasoning shortcuts and other artifacts to find the correct answers and obtain high scores~\cite{min2019compositional,chen2019understanding,Trivedi2020DiRe}. Such shortcuts arise from various factors, such as overly specific sub-questions, train-test leakage, and insufficient distractors.  These factors allow models to circumvent \emph{connected reasoning}---they need not read the context to find answers to previous sub-question(s) or use these answers to answer the later sub-questions that depend on them.

\added{The left hand side of Fig.~\ref{fig:introduction} illustrates an instance of this problem in an actual question (\leftquestion) taken from the \hpqa dataset~\cite{hotpotqa}. This question has the over-specification issue. At first glance, it appears to require a model to identify \emph{Kurt Vonnegut} as the author of \emph{Armageddon in Retrospect}, and then use this information to answer the final question about the famous satire novel he authored. However, this framing of the question is insufficient to enforce that models must perform connected \mh reasoning to arrive at the correct answer. A model can, in fact, find the correct answer to this question from the context without finding the answer to Q1. This is because, even if a model does not know that A1 refers to \textit{Kurt Vonnegut}, there happens to be only one person best known for a \textit{satirical novel} mentioned in the context.}

\added{Contrast this with the question on the right (\rightquestion), which cannot be answered by simply returning a novel that \textit{someone} was best known for. There are three possible answers in the context and choosing between them requires knowing which author is referenced. This is a desirable \mh question that requires connected reasoning.}

Prior work has characterized such reasoning, where a model arrives at the correct answer without using all supporting facts, as Disconnected Reasoning~\cite{Trivedi2020DiRe}. While this characterization enables filtering or automatically transforming existing datasets~\cite{Trivedi2020DiRe}, we ask a different question: \emph{How can we construct a new \mh dataset that, by design, enforces connected reasoning?}

We make two main contributions towards this:

\textbf{1) A new dataset construction approach:}
We introduce a bottom-up process for building challenging \mh reading comprehension QA datasets by carefully selecting and composing \sh questions obtained from existing datasets. The key ideas behind our approach are: (i) Composing multihop questions from a large collection of \sh questions, which allows a systematic exploration of a vast space of candidate \mh questions. (ii) Applying a stringent set of filters that ensure no sub-question can be answered without finding the answer to the previous sub-questions it is connected to \added{(a key property we formally define as part of the \textbf{\condition condition}, Eqn.~(\ref{eqn:musique-condition}))}. (iii) Reducing train-test leakage at the level of each single-hop question, thereby mitigating the impact of simple memorization tricks. (iv) Adding distractor contexts that cannot be easily identified. (v) Creating unanswerable \mh questions at the sub-question level.
 
\textbf{2) A new challenge dataset and empirical analysis:}
We build a new \mh QA dataset, \dataans (abbreviated as \ma), with $\sim$25K 2-4 hop questions with six different composition structures (cf.~Table~\ref{table:musiq-examples}). We demonstrate that \ma is more challenging and less cheatable than two prior \mh reasoning datasets, \hpqa~\cite{hotpotqa} and \twowikifull~\cite{xanh2020_2wikimultihop}. In particular, it has 3x the human-machine gap, and a substantially lower disconnected reasoning (DiRe) score which captures the extent to which a dataset can be cheated via disconnected reasoning~\cite{Trivedi2020DiRe}. We also show how various features of our dataset construction pipeline help increase dataset difficulty and reduce cheatability. Lastly, by incorporating the notion of insufficient context~\cite{rajpurkar2018know,Trivedi2020DiRe}, we also release a variant of our dataset, \mf, having $\sim$50K \mh questions which form contrasting pairs~\cite{kaushik2019learning,gardner2020evaluating} of answerable and unanswerable questions. \mf is even more challenging and harder to cheat on.

We hope our bottom-up \mh dataset construction methodology and our challenging datasets with a mixed number of hops will help develop proper \mh reasoning systems and decomposition-based models.

\section{Related Work}

\paragraph{Multihop QA.}
\ma is closest to HotpotQA \cite{hotpotqa} and 2WikiMultihopQA~\cite{xanh2020_2wikimultihop}. \hpqa was constructed by directly crowdsourcing 2-hop questions without considering the difficulty of composition and has been shown to be largely solvable without \mh reasoning~\cite{min2019compositional,chen2019understanding,Trivedi2020DiRe}. While \twowikifull was also constructed via composition, they use a limited set of hand-authored compositional rules, making it easy for large language models. We show that \ma is harder and less cheatable than both of these. Other \mh datasets~\cite[][\textit{inter alia}]{MultiRC2018,Dua2019DROP} focus on different challenges such as multiple modalitites~\cite{chen2020hybridqa,talmor2021multimodalqa}, open-domain QA~\cite{geva2021strategyqa,khot2020qasc}, fact verification~\cite{jiang2020hover}, science explanations~\cite{jansen-etal-2018-worldtree}, and relation extraction~\cite{welbl-etal-2018-constructing}, among others. Extending our ideas to these challenges is an interesting avenue for future work.

\paragraph{Unanswerable QA.}
Prior works have used unanswerable questions for robust reasoning in \sh~\cite{rajpurkar2018know} and \mh~\cite{ferguson-etal-2020-iirc,Trivedi2020DiRe} settings. IIRC~\cite{ferguson-etal-2020-iirc} focuses on open-domain QA where the unanswerable questions are identified by crowdsourcing questions where relevant knowledge couldn't be retrieved from Wikipedia. Our idea to make unanswerable \mh questions by removing support paragraphs is most similar to \citet{Trivedi2020DiRe}. While they rely on annotations (potentially incomplete) to identify these support paragraphs, we can use the bridge entities to remove any potential support paragraphs (containing the bridge entity) and better ensure unanswerability.
  
\paragraph{Question Decomposition and Composition.}
\Mh QA datasets have been decomposed into simpler questions~\cite{decomprc, talmor-berant-2018-web} and special meaning representations~\cite{Wolfson2020Break}. Our dataset creation pipeline naturally provides question decompositions, which can can help develop interpretable models~\cite{decomprc,khot-etal-2021-text}.

Recent work has also used bottom-up approaches to create \mh questions~\cite{pan-etal-2021-unsupervised,yoran2021turning} using rule-based methods. However, \emph{their primary goal was data augmentation} to improve on downstream datasets. The questions themselves haven't been shown to be challenging or less cheatable.

\begin{table*}[ht!]
    \scriptsize
    \centering
    \renewcommand{\arraystretch}{0}
    \begin{tabular}{
        >{\centering\arraybackslash}p{1.6cm}
        p{4.8cm}p{8.2cm}
    }

    \toprule
        

        Graph & Question & Decomposition \\
        \midrule

        \raisebox{-0.8\height}{\includegraphics[height=0.25\linewidth]{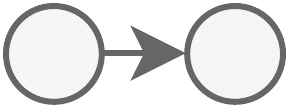}} &
        Who succeeded the first President of Namibia? \bridgetwo{Hifikepunye Pohamba} &

        \textbf{1.} Who was the first President of Namibia? \bridgeone{Sam Nujoma} \newline
        \textbf{2.} Who succeeded \bridgeone{Sam Nujoma}? \bridgetwo{Hifikepunye Pohamba} 
        \\
        \midrule

        \raisebox{-0.8\height}{\includegraphics[height=0.25\linewidth]{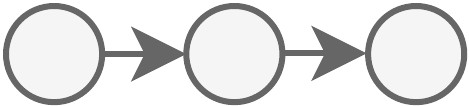}} &
        What currency is used where Billy Giles died? \bridgethree{pound sterling} &

        \textbf{1.} At what location did Billy Giles die? \bridgeone{Belfast} \newline
        \textbf{2.} What part of the UK is \bridgeone{Belfast} located in? \bridgetwo{Northern Ireland} \newline
        \textbf{3.} What is the unit of currency in \bridgetwo{Northern Ireland}? \bridgethree{pound sterling} 
        \\
        \midrule

        \raisebox{-0.8\height}{\includegraphics[height=0.6\linewidth]{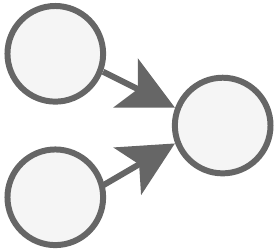}} &
        When was the first establishment that McDonaldization is named after, open in the country Horndean is located? \bridgethree{1974} &

        \textbf{1.} What is McDonaldization named after? \bridgeone{McDonald's} \newline
        \textbf{2.} Which state is Horndean located in? \bridgetwo{England} \newline
        \textbf{3.} When did the first \bridgeone{McDonald}’s open in \bridgetwo{England}? \bridgethree{1974} 
        \\
        \midrule

        \raisebox{-0.8\height}{\includegraphics[height=0.65\linewidth]{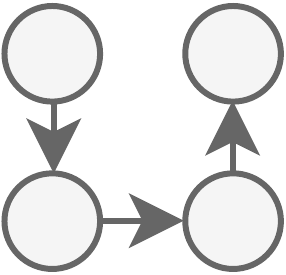}} &
        When did Napoleon occupy the city where the mother of the woman who brought Louis XVI style to the court died? \bridgefour{1805} &

        \textbf{1.} Who brought Louis XVI style to the court? \bridgeone{Marie Antoinette} \newline
        \textbf{2.} Who’s mother of \bridgeone{Marie Antoinette}? \bridgetwo{Maria Theresa} \newline
        \textbf{3.} In what city did \bridgetwo{Maria Theresa} die? \bridgethree{Vienna} \newline
        \textbf{4.} When did Napoleon occupy \bridgethree{Vienna}? \bridgefour{1805} 
        \\
        \midrule

        \raisebox{-0.8\height}{\includegraphics[height=0.6\linewidth]{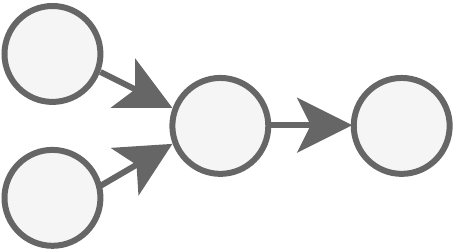}} &
        How many Germans live in the colonial holding in Aruba's continent that was governed by Prazeres's country? \bridgefour{5 million} &

        \textbf{1.} What continent is Aruba in? \bridgeone{South America} \newline
        \textbf{2.} What country is Prazeres? \bridgetwo{Portugal} \newline
        \textbf{3.} Colonial holding in \bridgeone{South America} governed by \bridgetwo{Portugal}? \bridgethree{Brazil} \newline
        \textbf{4.} How many Germans live in \bridgethree{Brazil}? \bridgefour{5 million} 
        \\
        \midrule

        \raisebox{-0.8\height}{\includegraphics[height=0.55\linewidth]{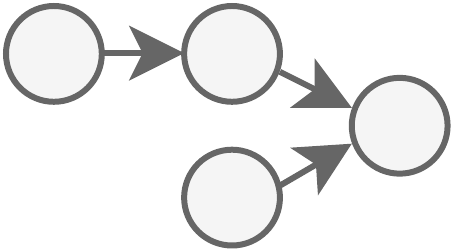}} &
        When did the people who captured Malakoff come to the region where Philipsburg is located? \bridgefour{1625} &

        \textbf{1.} What is Philipsburg capital of? \bridgeone{Saint Martin} \newline
        \textbf{2.} \bridgeone{Saint Martin} is located on what terrain feature? \bridgetwo{Caribbean} \newline
        \textbf{3.} Who captured Malakoff? \bridgethree{French} \newline
        \textbf{4.} When did the \bridgethree{French} come to the \bridgetwo{Caribbean}? \bridgefour{1625} 
        \\

        \bottomrule
    \end{tabular}
    \caption{
    \taclvone{\data has questions for 6 reasoning graphs (top). We show examples of 3 of them (bottom). An extended version of this table can be found in the appendix (Table ~\ref{table:musiq-extended-examples})}
    \added{The six reasoning graph shapes (2-hop to 4-hop) present in \data, along with sample questions.}
    }

    \label{table:musiq-examples}
\end{table*}

\section{Multihop Reasoning Desiderata}
\label{sec:multihop-desiderata}

\Mh question answering can be seen as a sequence of inter-dependent reasoning steps leading to the answer. In its most general form, these reasoning steps and the dependencies can be viewed as directed acyclic graph (DAG), $G_Q$. Each node $q_i$ in this graph represents a reasoning step or a ``hop'', e.g., a single-hop question in \mh QA or a KB relation traversal in graph-based KBQA. An edge $(q_j, q_i) \in \textrm{edges}(G_Q)$ indicates that the reasoning step $q_i$ relies critically on the output of the predecessor step $q_j$. For example, in Fig.~\ref{fig:introduction}, the single-hop question $Q2'$ depends on the answer to $Q1'$, and the graph $G_{Q'}$ is a linear chain $Q1' \rightarrow Q2'$.

Given this framing, a key desirable property for \mh reasoning is \textbf{connected reasoning}: \emph{Performing each step $q_i$ correctly should require the output of all its predecessor steps $q_j$}.

\textbf{Analytical Intuition:}
Suppose a model $M$ can answer each $q_i$ correctly with probability $p$, and it can also answer $q_i$ \emph{without the output of all its predecessor steps} with probability $r \leq p$. For simplicity, we assume these probabilities are independent across various $q_i$. $M$ can correctly answer a $k$-hop question $Q$ by identifying and performing all its $k$ reasoning steps. This will succeed with probability at most $p^k$. Alternatively, as an extreme case, it can ``cheat'' by identifying and performing only the last step $q_k$ (the ``end question'') without considering the output of $q_{k-1}$ (or other steps) at all. This could succeed with probability as much as $r$, which \emph{does not decrease with $k$} and is thus undesirable when constructing \mh datasets. Our goal is to create \mh questions that enforce connected reasoning, i.e., where $r \ll p$ and, in particular, $r < p^k$, so that models have an incentive to perform all $k$ reasoning steps.

Not surprisingly, the connected reasoning property is often not satisfied by existing datasets~\cite{min2019compositional,chen2019understanding,Trivedi2020DiRe}, and never optimized for during dataset construction. As a consequence, models are able to exploit artifacts in existing datasets that allow them to achieve high scores while bypassing some of the reasoning steps, thus negating the main purpose of building \mh datasets. Prior work~\cite{Trivedi2020DiRe} has attempted to measure the extent of connected reasoning in current models and datasets. However, due to the design of existing datasets, this approach is only able to measure this by ablating the pre-requisites of each reasoning step, i.e., the supporting facts. Rather than only measure, we propose a method to \emph{construct} \mh QA datasets that directly optimize for this condition.

Consider question \leftquestion on the left hand side of Fig.~\ref{fig:introduction}. It can be answered in two steps, Q1 and Q2. However, the information in Q2 itself is sufficient to uniquely identify A2 from the context, \textit{even without considering} A1. That is, while there is an intended dependency between Q1 and Q2, Q2 can be answered correctly \emph{without} requiring the output of its predecessor question Q1. Our approach constructs \mh questions that prevent this issue, and thereby require the desired connected reasoning. \added{Specifically, we carefully choose which \sh questions to compose and what context to use such that each constituent \sh question necessitates the answers from one or more previous questions.}

\section{Connected Reasoning via Composition}
\label{sec:multihop-composition}

The central issue we want to address is ensuring connected reasoning. Our solution is to use a bottom-up approach where we compose multihop questions from a large pool of single-hop questions. As we show later, this approach allows us to explore a large space of \mh questions and carefully select ones that require connected reasoning. Additionally, with each \mh question, we will have associated constituent questions, their answers and supporting paragraphs, which can help develop more interpretable models. Here we describe the high-level process and describe the specifics in the next section.

\subsection{Multihop via Single-hop Composition}
As mentioned earlier, \mh questions can be viewed as a sequence of reasoning steps where answer from one reasoning step is used to identify the next reasoning step. Therefore, we can use \sh questions containing answers from other questions to construct potential \mh questions. E.g., in Fig.~\ref{fig:introduction}, Q2' mentions A1', and hence \sh questions Q1' and Q2' can be composed to create a DAG $Q1' \rightarrow Q2'$ and \mh question \rightquestion (right). Concretely, to create a \mh question from two \sh questions, we have a \textbf{composability criteria}: Two single-hop question answer tuples $(q_1, a_1)$ and $(q_2, a_2)$ are composable into a multihop question $Q$ with $a_2$ as a valid answer if $a_1$ is a named entity and it is mentioned in $q_2$. See \S\ref{sec:construction-pipeline}:S2 for detailed criteria.

This process of composing multihop questions can be chained together to form candidate reasoning graphs of various shapes and sizes (examples in Table~\ref{table:musiq-examples}). Formally, each \mh question $Q$ has an underlying DAG $G_Q$ representing the composition of the \sh questions $q_1, q_2, \ldots, q_n$, which form the nodes of $G_Q$. A directed edge $(q_j, q_i)$ indicates that $q_i$ depends on the answer of the previous sub-question $q_j$. $a_i$ is the answer to $q_i$, and thereby, $a_n$ is the answer to $Q$.

\subsection{Ensuring Connected Reasoning}

Given the graph $G_Q$ associated with a question $Q$, ensuring connected reasoning requires ensuring that for each edge $(q_j, q_i) \in \textrm{edges}(G_Q)$, arriving at answer $a_i$ using $q_i$, necessitates the use of $a_j$. In other words, without $a_j$, there isn't sufficient information in $q_i$ to arrive at $a_i$.

The existence of such information can be probed by training a strong QA model $M$ on subquestions ($q_i$) with the mention of their predecessor's answer ($a_j$) masked out (removed). If, on held out data, the model can identify a subquestion's answer ($a_i$) without its predecessor's answer ($a_j$), we say the edge $(q_j, q_i)$ is disconnected. Formally, we say $Q$ \emph{requires connected reasoning} if:
\begin{align}
    \forall (q_j,q_i) \in \text{edges}(G_Q): M(\msinglemaskedqn{i}{j}) \ne a_i
\label{eqn:hop-connected}
\end{align}
where \singlemaskedqn{i}{j} denotes the subquestion formed from $q_i$ by masking out the mention of the answer $a_j$.

Consider the masked questions Q2 and Q2' in Fig.~\ref{fig:introduction}. While Q2 can easily be answered without answer A1, Q2' can't be answered without A1' and \rightquestion hence satisfies condition~(\ref{eqn:hop-connected}).

\subsection{Reading Comprehension Setting}
While our proposed framework makes no assumptions about the choice of the model, and is applicable to open-domain setting, we focus on the Reading Comprehension (RC) setting, where we've a fixed set of paragraphs as context, $C$.

In a RC setting, apart from requiring the dependence between the reasoning steps, we also want the model to depend on the context to answer each question. While this requirement seems unnecessary, previous works have shown that RC datasets often have artifacts that allow models to predict the answer without the context~\cite{kaushik-lipton-2018-much} and can even memorize the answers~\cite{lewis-etal-2021-question} due to train-test leakage. As we will show later, previous \mh RC datasets can be cheated via such shortcuts. To ensure the dependence between the question and context, we modify the required condition in Eqn.~(\ref{eqn:hop-connected}) to:
\begin{align}
\forall (q_j,q_i) \in \text{edges}(G_Q): M(\msinglemaskedqn{i}{j}; C) \ne a_i \nonumber\\
\wedge \quad \forall q_i \in \text{nodes}(G_Q): M(q_i; \phi) \ne a_i 
\label{eqn:musique-condition}
\end{align}

In summary, we want \mh reading comprehension questions that satisfy condition~(\ref{eqn:musique-condition}) for a strong trained model $M$. If it does, we say that the question satisfies the \textbf{\condition condition}. Our dataset construction pipeline optimizes for this condition as described next.

\begin{figure*}
    \centering
	\includegraphics[width=\textwidth]{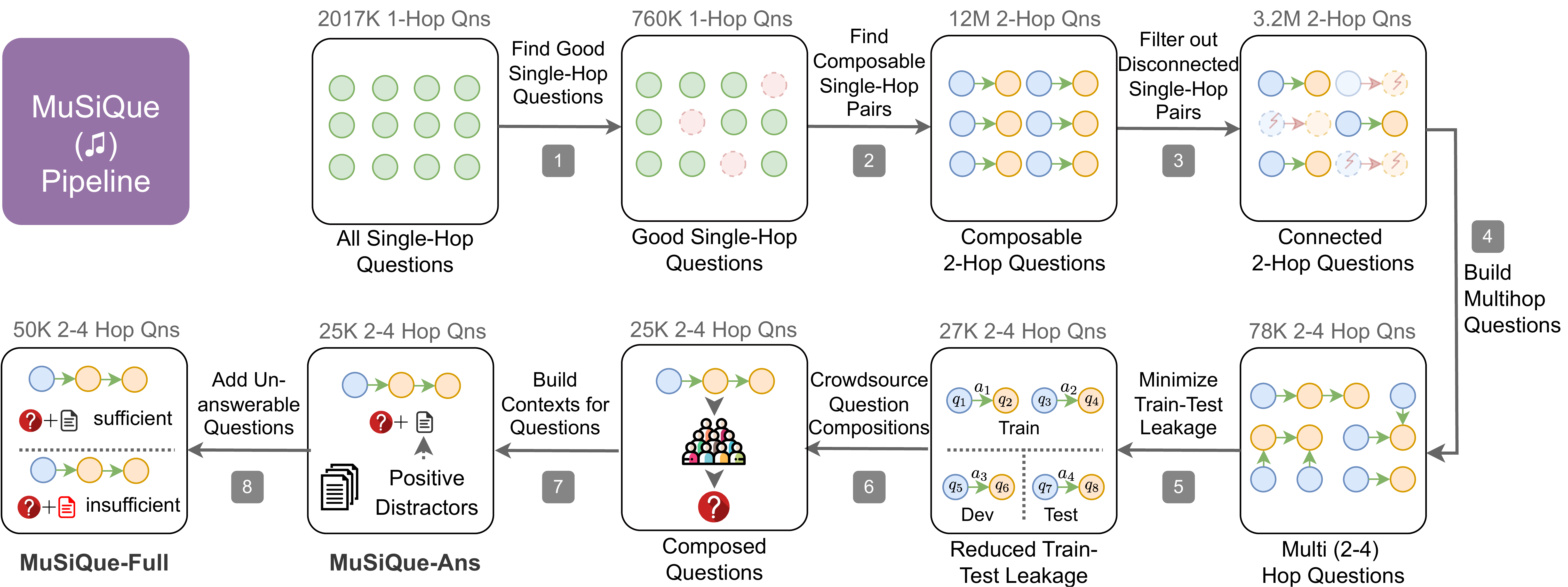}
	\caption{\data construction pipeline. \data pipline takes \sh questions from existing datasets, explores the space of \mh questions that can be composed from them, and generates dataset of challenging \mh questions that are difficult to cheat on. \data pipeline also makes unanswerable \mh questions that makes the final dataset significantly more challenging.}
	\label{fig:musiq-pipeline}
\end{figure*}

\section{Dataset Construction Pipeline}
\label{sec:construction-pipeline}

The high-level schematic of the pipeline is shown in Fig.~\ref{fig:musiq-pipeline}. We begin with a large set of RC \sh questions from 5 English Wikipedia-based datasets, SQuAD \cite{rajpurkar-etal-2016-squad}, Natural Questions~\cite{kwiatkowski-etal-2019-natural}, MLQA (en-en)~\cite{lewis2020mlqa}, T-REx~\cite{elsahar-etal-2018-rex}, Zero Shot RE \cite{levy-etal-2017-zero}, where instances are of the form $(q_i, p_i, a_i)$ referring to the question, the associated paragraph, and the answer, respectively. \added{For Natural Questions, as the context is very long (entire Wikipedia page), we use the annotated long answer (usually a paragraph) from the dataset as the context, and the annotated short answer as the answer.} Then, we take the following two steps:

\paragraph{S1. Find Good \SH Questions.}

Even a tolerably small percentage of issues in \sh questions can compound into an intolerably large percentage in the composed \mh questions. To mitigate this, we first remove questions that are likely annotation errors. Since manually identifying such questions at scale is laborious, we use a model-based approach. We remove the questions for which none of five large trained QA models\footnote{Two random-seed variants of RoBERTa-large~\cite{liu2019roberta}, two random-seeds of Longformer-Large~\cite{Beltagy2020Longformer} and one UnifiedQA~\cite{2020unifiedqa}.} can predict the associated answer with $> 0$ answer F1. Furthermore, we remove (i) erroneous questions where the answer spans are not in the context, (ii) questions with $<$ 20 word context as we found them to be too easy, and (iii) questions with $>$ 300 word context to prevent final \mh question context from being too long for current long-range transformer models.

\paragraph{S2. Find Composable \SH Pairs.}

To create 2-hop questions, we first collect distinct \sh question pairs with a bridge entity. Specifically, we find pairs ($q_1$, $p_1$, $a_1$) and ($q_2$, $p_2$, $a_2$) such that (i) $a_1$ is a named entity also mentioned in $q_2$, (ii) $a_2$ is not in $q_1$, and (iii) $p_1 \ne p_2$. Such pairs can be combined to form a 2-hop question ($Q$, \{$p_1$, $p_2$\}, $a_2$). To ensure the mentions ($a_1$ and its occurrence in $q_2$ denoted $e_2$) refer to the same entity, we ensure: \textbf{1.} Spacy entity tagger~\cite{spacy} tags $a_1$ and $e_2$ as entities of the same type. \textbf{2.} A Wikipedia search with $a_1$ and $e_2$ returns identical 1st result. \textbf{3.} A SOTA Wikification model~\cite{wu2019zero} returns the same result for $a_1$ and $e_2$. At a later step (S7) when humans write composed questions from DAGs, they get to remove questions containing erroneous pairs. Only 8\% of the pairs are pruned in that step, indicating that step S2 is quite effective.

\paragraph{S3. Filter Disconnected \SH Pairs.} 

We want connected 2-hop questions -- questions that cannot be answered without using the answers of the constituent \sh questions. The \condition condition (\ref{eqn:musique-condition}) states that for a 2-hop question to be connected, either sub-question $q_i$ should not be correctly answered without its context ($M(\munmaskedqn{i}, \phi) \neq a_i$) and the tail question $q_2$ should not be correctly answered when $a_1$ is removed from it ($M(\msinglemaskedqn{2}{1}, C) \neq a_2$). Accordingly we use a two step filtering process to find connected 2-hop questions. For simplicity, and because the second condition already filters some tail questions, our current implementation enforces the first condition only on the head question, $q_1$.

\textbf{Filtering Head Nodes}: 
We collect all questions that appear at least once as the head of composable 2-hop questions ($q_1$) to create a set of head nodes. 
We create 5-fold train-test splits of this set and train two Longformer-Large models (different seeds) per split (train on three, validate and test on one). We generate answer predictions using the 2 models on their corresponding test splits resulting in 2 predictions per question. 
We accept a head question if, on average, the predicted answers' word overlap (computed using answer f1) with the answer label is $<$ 0.5.

\textbf{Filtering Tail Nodes}: We create a unique set of masked \sh questions that occur as a tail node ($q_2$) in any composable 2-hop question. If the same \sh question occurs in two 2-hop questions with different masked entities, they both are added to the set. We combine the gold-paragraph with 9 distractor paragraphs (retrieved~\footnote{We use the BM25 algorithm via Elasticsearch.} using the question without the masked entities as query). As before, we create 5-fold train-test splits and use 2 Longformer-Large models to get 2 answer and support predictions. We accept a tail question if either mean answer F1 $\le 0.25$, or if it's $\le 0.75$ and mean support F1 $< 1.0$.

The thresholds for head and tail node filtering were chosen via a manual inspection of a few predictions in various ranges of the parameters, and gauging at what F1 values does the model's answer semantically match the correct answer (e.g., "Barack Obama" and "President Barack Obama" overlap with 0.8 answer F1). Controlling these thresholds provides a way to trade-off between the degree of cheatability allowed in the dataset and the size of the final dataset. We aim to limit cheatability while retaining a reasonable dataset size.
 
Finally, only 2-hop questions for which both head and tail node are acceptable are kept. We call this process \textbf{Disconnection Filtering}.

\paragraph{S4. Build Multihop Questions.}
We now have a set of connected 2-hop questions, which form directed edges of a graph. Any subset DAG of it can be used to create a connected \mh question. We use 6 types of reasoning graphs with 2-4 hops as shown in Table ~\ref{table:musiq-examples}. To avoid very long questions, we limit \sh questions to $\le 10$ tokens, the total length of questions in 2,3-hops to $\le 15$, and 3-hops to $\le 20$ tokens. To ensure diversity, we (1) cap the reuse of bridging entities and \sh questions at 25 and 100 \mh questions respectively (2) remove any n-hop question that's subset of any m-hop question (m $>$ n $>$ 1).

\paragraph{S5. Minimize Train-Test Leakage.}
We devise a procedure to create train, validation and test splits such that models cannot achieve high-scores via memorization enabled by train-test leakage, an issue observed in some existing datasets~\cite{lewis-etal-2021-question}. Our procedure ensures that the training set has no \textit{overlap} with validation or the test sets, and tries to keep the \textit{overlap} between validation and test sets minimal.

We consider two \mh questions $Q_i$ and $Q_j$ to \textit{overlap} if any of the following are common between $Q_i$ and $Q_j$: (i) \sh question (ii) answer to any \sh question (iii) associated paragraph to any \sh question. To minimize such \emph{overlap}, we take a set of \mh questions, greedily find a subset of given size (S) which least \emph{overlaps} with its complement (S'), and then remove \emph{overlapping} questions from S', to get train (S) and dev+test set (S'). Then, we split dev+test to dev and test similarly. We ensure the distribution of source datasets of \sh questions in train, dev and test are similar, and also control the proportion of 2-4 hop questions.

\paragraph{S6. Build Contexts for Questions.}

For an n-hop question, the context has 20 paragraphs containing: (i) supporting paragraphs associated with its \sh questions \{$p_1$, $p_2$ $\ldots$ $p_n$\}, (ii) distractor paragraphs retrieved using a query that is a concatenation of \sh questions from which all intermediate answer mentions are removed. To make distractor paragraphs harder to identify, we retrieve them from the set of gold-paragraphs for the filtered \sh question (S1). 

\paragraph{S7. Crowdsource Question Compositions.}

\added{We crowdsource question compositions on Amazon MTurk, where workers composed coherent questions from our final DAGs of \sh questions.
In the interface (Fig. ~\ref{fig:composition-annotation-interface}), workers could see a list of \sh questions with their associated paragraphs and how they are connected via bridge entities. They were first asked to check whether all pairs of mentions of bridge entities indeed refer to the same underlying entity. If they answered `yes' for each pair,\footnote{They answered yes 92\% of the time, on average.} they were asked to compose a natural language question ensuring that information from all \sh questions in the DAG is used, and the answer to the composed question is the same as the last \sh question. If they answered `no' for any of the pairs, we discarded that question. Our tutorial provided them with several handwritten good and bad examples for each of the 2-4 hop compositions. Workers were encouraged to write short questions and make implicit inferences when possible. They were allowed to split questions into two sentences if needed.}

\added{We carried out a qualification round where 100 workers participated to perform the aforementioned task on 20 examples each. We manually evaluated these annotations for correctness and coherence, and selected 17 workers to annotate the full dataset. To ensure dataset quality, we carried out crowdsourcing in 9 batches, reading 10-20 random examples from each worker after each batch and sending relevant feedback via email, if needed. Workers were paid 25, 40, and 60 cents for each 2, 3, and 4 hop question, amounting to $\sim$15 USD per hour, totaling $\sim$11K USD.}

We refer to the dataset at this stage as \textbf{\dataans} or \ma.

\paragraph{S8. Add Unanswerable Questions.}

For each answerable \mh RC instance we create a corresponding unanswerable \mh RC instance using the procedure similar to the one proposed in \cite{Trivedi2020DiRe}. For a \mh question we randomly sample any of its \sh question and make it unanswerable by ensuring the answer to that \sh question doesn't appear in any of the paragraphs in context (except this requirement, the context is built as described in S6). Since one of the \sh questions is unanswerable, the whole \mh question is unanswerable. 

The task now is to predict whether the question is answerable, and predict the answer and support if it's answerable. Given the questions for answerable and unanswerable pair are identical and the context marginally changes, models that rely on shortcuts find this new task very difficult. We call the dataset at this stage \textbf{\datafull} or \mf, and both datasets together as \textbf{\data}.

\paragraph{Final Dataset.}

\begin{table}[t]
    \centering
    \small
    \begin{tabular}{ccccc}\toprule

                & 2-hop & 3-hop & 4-hop & Total (24,814) \\
      \midrule

      Train     & 14376 & 4387  & 1175  & 19938 \\
      Dev       & 1252  & 760   & 405   & 2417  \\
      Test      & 1271  & 763   & 425   & 2459  \\

      \bottomrule
    \end{tabular}
    \caption{Dataset statistics of \dataans. \datafull contains twice the number of questions in each category above -- one answerable and one unanswerable.}
    \label{table:musiq-stats}
\end{table}

The statistics for \ma (\mf has twice the number of questions in each cell) are shown in Table~\ref{table:musiq-stats}. \data constitutes unique 21020 \sh questions, 4132 answers to \mh questions, 19841 answers to \sh questions, and 7676 supporting paragraphs. \data has 6 types of reasoning graphs and 2-4 hops (cf.~Table~\ref{table:musiq-examples} for examples). \\

In summary, our construction pipeline allows us to produce a dataset with mixed hops, multiple types of reasoning graphs, and unanswerable sub-questions, all of which make for a more challenging and less cheatable dataset (as we will quantify in Section~\ref{sec:exp-results}). Question decomposition, which is a natural outcome of our construction pipeline, can also be used to aid decomposition-based QA research~\cite{decomprc,khot-etal-2021-text}.

\section{Dataset Quality Assessment}
\label{sec:dataset-quality}

\paragraph{Quality of \ma.}
To assess the quality of \ma, we first evaluate how well humans can answer questions in it. Note that we already have gold answers and supporting paragraphs from our construction pipeline. This goal is therefore not to determine gold labels, but rather to measure how well humans perform on the task treating our gold labels as correct.

We sample 125 questions from \ma validation and test sets, and obtain 3 annotations (answer and supporting paragraphs) for each question. We used Amazon Mechanical Turk,\footnote{\url{https://www.mturk.com}} selecting crowdsource workers as described in \S\ref{subsubsec:human-performance}.

Workers were shown the question and all paragraphs in the context, and were asked to highlight the answer span and checkmark the supporting paragraphs. Our interface (Fig. ~\ref{fig:validation-annotation-interface}) allowed for searching, sorting, and filtering the list of paragraphs easily with interactive text-overlap-based search queries. The instructions included worked out examples. 

We compute human performance by comparing against gold labels for answer and support in two ways: 
1) \textbf{Human Score}---the most frequent answer and support among the three annotators \added{breaking ties at random} (the strategy used by \citet{rajpurkar2018know}), and 2) \textbf{Human Upper Bound (UB)}---the answer and support that maximizes the score (as done by \citet{hotpotqa}).

Furthermore, to assess how well human agree with each other (ignoring our gold labels), we also compute the \textbf{Human Agreement (Agr)} score~\cite{rajpurkar-etal-2016-squad,hotpotqa}. Specifically, we treat one of 3 annotations, chosen randomly, as predicted, and evaluate it against rest of the annotations, which are treated as correct.

\begin{table}[htb]
    \centering
    \small
    \setlength{\tabcolsep}{5pt}
    \begin{tabular}{
        >{\centering\arraybackslash}p{1.6cm}
        >{\centering\arraybackslash}p{1.2cm}
        >{\centering\arraybackslash}p{1.2cm}
        >{\centering\arraybackslash}p{1.2cm}
    }\toprule
        Human & Score    & UB     & Agr \\
        \midrule
        Answer F1          & 78.0        & 88.6    & 84.1 \\
        Support F1         & 93.9        & 97.3    & 91.4 \\
        \bottomrule
    \end{tabular}
    \caption{Human performance (score and upper bound) and agreement on \dataans.}
    \label{table:musiq-human-scores}
\end{table}

Table~\ref{table:musiq-human-scores} demonstrates that \ma is a high-quality dataset. Furthermore, as we will discuss in \S\ref{subsubsec:human-performance}, we also compare our human performance with two other similar datasets (HotpotQA and 2WikiMultihopQA), and show that \ma is close to them under these metrics (\S\ref{sec:exp-results}).

\paragraph{Quality of \mf.}
We perform an additional manual validation to assess dataset quality of \mf. Recall that \mf shares the answerable questions with \ma, the only extra task in \mf being determining the answerability of a question from the given context. To assess the validity of this task, we sampled 50 random instances from \mf, and one of the authors determined the answerability of each question from its context. We found that in 45 out of the 50 instances (90\%) the human predicted answerability matched the gold label, showing that \mf is a also high-quality dataset.

\paragraph{Multihop Nature of \data.}

Finally, we assess the extent to which \ma satisfies the \textbf{MuSiQue condition} (Eqn.~\ref{eqn:musique-condition}) for connected reasoning. To this end, we first estimate what percentage of head and tail questions in the validation set would we retain if we were to repeat our disconnection filtering procedure (S3) with models trained on the final training data. This captures the fraction of the questions in \ma that satisfy the MuSiQue condition. We then compare it with the respective numbers from the original step S3.

In the original disconnection filtering step, we retained only 26.5\% of the tail questions, whereas we would have retained 79.0\% of the tail questions had we filtered the final validation dataset. For head questions, we see a less dramatic but still significant effect---we originally retained 74.5\% questions, and would now have retained 87.7\% had we filtered the final validation set. This shows that vastly more questions in \ma satisfy the MuSiQue condition than what we started with.

\section{Experimental Setup} 
\label{sec:exp-setup}

\subsection{Datasets}
\label{subsubsec:datasets}

We compare our datasets (\dataans and \datafull) with two similar \mh RC datasets: distractor-setting of \hpqa~\cite{hotpotqa} and \twowikifull~\cite{xanh2020_2wikimultihop}.\footnote{For brevity, we use \hq, \tw, \ma/Full to refer to \hpqa, \twowikifull, \dataans/Full, resp.} Both datasets have 10 paragraphs as context. \hq and \tw have 2-hop and 2,4-hop questions respectively. Additionally, \hq has sentence support and \tw has entity-relation tuples support, but we don't use this annotation in our training or evaluation for a fair comparison.

\hq, \tw, and \ma have 90K, 167K, and 20K training instances, respectively. For a fair comparison, we use equal sized training sets in all our experiments, obtained by randomly sampling 20K instances each from \hq and \tw, and referred to as \hq-20k and \tw-20k, respectively.

\paragraph{Notation.} Instances in \ma, \hq, and \tw are of the form $(Q, C; A, P_s)$. Given a question $Q$ and context C consisting of a set of paragraphs, the task is to predict the answer $A$ and identify supporting paragraphs $P_s \in C$. \ma additionally has gold decomposition $G_Q$ (\S\ref{sec:multihop-desiderata}), which can be leveraged during training. Instances in \mf are of form $(Q, C; A, P_s, S)$, where there's an additional binary classification task to predict $S$, the answerability of $Q$ based on $C$, also referred to as context \emph{sufficiency}~\cite{Trivedi2020DiRe}.

\paragraph{Metrics.} For \ma, \hq, and \tw, we report the standard F1 based metrics for answer (\textbf{An}) and support identification (\textbf{Sp}); see \citet{hotpotqa} for details. 
To make a fair comparison across datasets, we use only paragraph-level support F1.

For \mf, we follow~\citet{Trivedi2020DiRe} to combine sufficiency prediction $S$ with An and Sp, which are denoted as An+Sf and Sp+Sf. Instances in \mf are evaluated in pairs. For each $Q$ with a sufficient context $C$, there is a paired instance with $Q$ and an insufficient context $C'$. For An+Sf, if a model incorrectly predicts context sufficiency (yes or no) for either of the instances in a pair, it gets 0 pts on that pair. Otherwise, it gets same An score on that pair as it gets on the answerable instance in that pair. Scores are averaged across all pairs of instances in the dataset. Likewise for Sp+Sf.

\subsection{Models}

Our models are Transformer-based~\cite{vaswani2017attention} language models~\cite{devlin-etal-2019-bert}, implemented using PyTorch~\cite{NEURIPS2019_9015}, HuggingFace Transformers~\cite{Wolf2019HuggingFacesTS} and AllenNLP~\cite{Gardner2017AllenNLP}. We experiment with 2 types of models: (1) \emph{\Mh Models}, which are in principle capable of employing desired reasoning, and have demonstrated competitive performance on previous \mh QA datasets. They help probe the extent to which a dataset can be solved by current models. (2) \emph{Artifact-based Models}, which are restricted in some way that prohibits them from doing desired reasoning (discussed shortly). They help probe the extent to which a dataset can be cheated. Next, we describe these models for \ma and \mf. For \hq and \tw, they work similar to \ma.

\subsubsection{\Mh Models}

\paragraph{End2End (EE) Model.} This model takes $(Q, C)$ as input, runs it through a transformer, and predicts $(A, P_s)$ as the output for \ma and $(A, P_{s}, S)$ for \mf. We use Longformer-Large as it's one of the few transformer architectures that is able to fit the full context, and follow ~\citet{Beltagy2020Longformer} for answer and support prediction. Answerability prediction is done via binary classification using \texttt{CLS} token.

Note that our Longformer EE model is a strong model for \mh reasoning. When trained on full datasets, its answer F1 is 78.4 (within 3 pts of published SOTA~\cite{groeneveld2020simple}) on \hq, and 87.7 (SOTA) on \tw.

\paragraph{Select+Answer (SA) Model.} 
This model, inspired by Quark~\cite{groeneveld2020simple} and SAE~\cite{tu2020select}, has two parts. First, a \textit{selector} ranks and selects the $K$ most relevant paragraphs $C_K \subseteq C$.\footnote{$K$ is a hyperparameter, chosen from \{3,5,7\}.} \added{Specifically, given $(Q, C)$ as input, it classifies every paragraph $P \in C$ as relevant or not, and is trained with the cross-entropy loss.} Second, for \dataans, the \textit{answerer} predicts the answer and supporting paragraphs based only on $C_K$. For \datafull, it additionally predicts answerability. Both components are trained individually using annotations available in the dataset. We implement a selector using RoBERTa-large~\cite{liu2019roberta}, and an answerer using Longformer-Large.

\paragraph{Step Execution (EX) Model.} Similar to prior work~\cite{talmor-berant-2018-web,decomprc,qi2021answering,khot-etal-2021-text}, this model performs explicit, step-by-step \mh reasoning, by first \textit{decomposing} the $Q$ into a DAG $G_Q$ having \sh questions, and then calling \sh model repeatedly to \emph{execute} this decomposition.

The \textit{decomposer} is trained with gold decompositions, and is implemented with BART-large.

The \textit{executor} takes $C$ and the predicted DAG $G_Q$, and outputs $(A, P_s)$ for \dataans and $(A, P_s, S)$ for \datafull. It calls \sh model $M_s$ repeatedly while traversing $G_Q$ along the edges and substituting the answers.

Model $M_s$ is trained on only \sh instances---taking $(q_i, C)$ as input, and producing $(A, P_i)$ or $(A, P_{s_i}, S_i)$ as the output. Here $P_i$ refers to the supporting paragraph for $q_i$ and $S_i$ refers to whether $C$ is sufficient to answer $q_i$. For \datafull, the answerer predicts $Q$ as having sufficient context if $M_s$ predicts all $q_i$ to have sufficient context. We implement 2 such \sh models $M_s$: End2End and Select+Answer, abbreviated as \textbf{EX(EE)} and \textbf{EX(SA)} respectively

We don't experiment with this model on \hq, since it needs ground-truth decomposition and intermediate answers, which aren't available in \hq.

\paragraph{Baseline (RNN) Model.} \added{The filtering steps in our pipeline use transformer-based models, which could make \data particularly difficult for transformer-based models. A natural question then is, can a strong non-transformer model perform better on \data? To answer this, we evaluate our re-implementation of a strong RNN-based baseline~\cite{hotpotqa} (see their original paper for details). To verify our implementation, we trained it on full \hpqa and found its performance to be 64.0 An (answer F1) on the validation set, better than what's reported by \citet{hotpotqa} (58.3 An). We thus use this model as a strong non-transformer baseline.}

\subsubsection{Artifact-based Models}
\label{subsubsec:partial-input-models}

The \textbf{Q-Only Model} takes only $Q$ as input (no $C$) and generates output $A$ for \ma and $(A, S)$ for \mf. We implement this with BART-large~\cite{lewis-etal-2020-bart}. The \textbf{C-Only Model} takes only $C$ as input (no $Q$) and predicts $(A, P_s)$ for \ma and $(A, P_s, S)$ for \mf. We implement this with an EE Longformer-Large model with empty $Q$. The \textbf{1-Para Model}, like \citet{min2019compositional,chen2019understanding}, is similar to SA model with $K$=1.
Instead of training the selector to rank all $P_s$ the highest, we train it to rank any paragraph containing the answer $A$ as the highest. The answerer then takes as input one selected paragraph $p \in P_s$ and predicts an answer to $Q$ based solely on $p$. This model can't access full supporting information as all considered datasets have at least 2 supporting paragraphs.

\subsubsection{Cheatability Score}

We compute the \textbf{DiRe score} of all datasets, which measures the extent to which the datasets can be cheated by strong models via Disconnected Reasoning~\cite{Trivedi2020DiRe}. We report scores based on the SA model since it performed the best.

\subsection{Human Performance}
\label{subsubsec:human-performance}

Apart from assessing the human performance level on \ma, as discussed in \S\ref{sec:dataset-quality}, we also obtain human performance on \hq and \tw. For a fair comparison, we use the same crowdsourcing workers, annotation guidelines, and interface across the 3 datasets. We sample 125 questions from each dataset, shuffle them all into one set, and obtain 3 annotations per question for answer and support.

To select the workers, we ran a qualification round where each worker was required to identify answer and support for at least 25 questions. We then selected workers who had more than 75 An and Sp scores on all datasets. 7 out of 15 workers were qualified for rest of the validation.

\section{Empirical Findings}
\label{sec:exp-results}

We now discuss our findings, demonstrating that \data is a challenging \mh dataset that is harder to cheat on than existing datasets (\S\ref{subsec:musiq-challenging}) and that the steps in the \data construction pipeline are individually valuable (\S\ref{subsec:pipeline-valuable}). Finally, we explore avenues for future work (\S\ref{subsec:misc-improvements}).

For \hq and \tw, we report validation set performance. For \ma and \mf, Table~\ref{table:musiq-full-better} reports test set numbers; all else is on the validation set.

\subsection{\data is a Challenging Dataset}
\label{subsec:musiq-challenging}

Compared to \hq and \tw, both variants of \data are less cheatable via shortcuts and have a larger human-to-model gap.

\begin{table}[ht]
    \centering
    \small
    \setlength{\tabcolsep}{4.5pt}
    \begin{tabular}{
        >{\centering\arraybackslash}p{0.5cm}
        >{\centering\arraybackslash}p{1.0cm}
        >{\centering\arraybackslash}p{1.5cm}
        >{\centering\arraybackslash}p{1.5cm}
        >{\centering\arraybackslash}p{1.5cm}
    }\toprule
                    &              & HQ-20K           & 2W-20K     & \iconm-Ans \\

                    &              & \p An $|$ Sp & \p An $|$ Sp & \p An $|$ Sp \\
        \midrule

        \multirow{2}{*}{\rotatebox[origin=c]{90}{\parbox[c]{0.9cm}{\centering Hu-man}}}
                    & Score          & 84.5 $|$ 92.5       & 83.2 $|$ 99.3    & 78.0 $|$ 93.9 \\
                    & UB           & 91.8 $|$ 96.0       & 89.0 $|$ 100\x   & 88.6 $|$ 97.3 \\

        \midrule
        \multirow{2}{*}{\rotatebox[origin=c]{90}{\parbox[c]{1.7cm}{\centering \Mh Models}}}
                   & RNN           & 51.0 $|$ 82.4       & 52.7 $|$ 94.9    & 13.6 $|$ 41.9 \\
                   & EE            & 72.9 $|$ 94.3       & 72.9 $|$ 97.6    & 42.3 $|$ 67.6 \\
                   & SA            & 74.9 $|$ 94.6       & 79.5 $|$ 99.0    & 47.3 $|$ 72.3 \\
                   & EX(EE)        &      \mpt           & 79.8 $|$ 97.5    & 45.6 $|$ 77.8 \\
                   & EX(SA)        &      \mpt           & 71.2 $|$ 98.1    & 49.8 $|$ 79.2 \\

        \midrule
        \multirow{2}{*}{\rotatebox[origin=c]{90}{\parbox[c]{1.2cm}{\centering Artifact Models}}}
                   & 1-Para        & 64.8 $|$ \mpt       & 60.1 $|$ \mpt    &  32.0 $|$ \mpt \\
                   & C-only        & 18.4 $|$ 67.6       & 50.1 $|$ 92.0    & \p3.4 $|$ \p0.0 \\
                   & Q-only        & 19.6 $|$ \mpt       & 27.0 $|$ \mpt    & \p4.6 $|$ \mpt \\

        \midrule
        \multicolumn{2}{c}{DiRe Score}
                   & 68.8 $|$ 93.0 & 63.4 $|$ 98.5       & 37.8 $|$ 63.4 \\
        \bottomrule
    \end{tabular}
    \caption{
        \added{Compared to the other datasets considered, \ma has a much larger human-model gap (higher gap between top and middle sections), and is much less cheatable (lower scores in bottom two sections).}
    }
    \label{table:musiq-vs-others}
\end{table}

\paragraph{Higher Human-Model Gap.}

Top two sections of Table~\ref{table:musiq-vs-others} show \ma has a significantly higher human-model gap \added{(computed as Human Score minus best model score)} than the other datasets, for both answer and supporting paragraph identification. In fact, for both the other datasets, supporting paragraph identification has even surpassed the human score, whereas for \ma, there is 14 pts gap. Additionally, \ma has a $\sim$27 pt gap in answer F1, whereas \hq and \tw have a gap of only 10 and 5, respectively.

Our best model, EX(SA), scores 57.9, 47.9, and 28.1 answer F1 on 2, 3, and 4-hop questions of \ma, resp. The EE model, on the other hand, stays around 42\% irrespective of the number of hops.

\paragraph{Lower Cheatability.}
The 3rd section of Table~\ref{table:musiq-vs-others} shows that the performance of artifact-based models (\S\ref{subsubsec:partial-input-models}) is much higher on \hq and \tw than on \ma. E.g., the 1-Para model achieves 64.8 and 60.1 answer score on \hq and \tw, resp., but only 32.0 on \ma. Support identification in both datasets can be done to a surprisingly high degree (67.6 and 92.0 F1) even without the question (C-only model), but fails on \ma.\footnote{Even when \ma is modified to have 10 paragraphs like \hq, C-only support score remains low; cf.~Table~\ref{table:positive-distractors-better}.}

Similarly, the last row of Table~\ref{table:musiq-vs-others} shows that the \textbf{DiRe} answer scores of \hq and \tw (68.8 and 63.4) are high, indicating that even disconnected reasoning \added{(bypassing reasoning steps)} can achieve such high scores. In contrast, this number is significantly lower (37.8) for \ma.

These results demonstrate that \ma is significantly less cheatable via shortcut-based reasoning.

\paragraph{\datafull: Even More Challenging.}

Table~\ref{table:musiq-full-better} shows that \mf is significantly more difficult and less cheatable than \ma.

\begin{table}[ht]
    \centering
    \small
    \begin{tabular}{
        >{\centering\arraybackslash}p{0.7cm}
        >{\centering\arraybackslash}p{1.4cm}
        ccc
    }\toprule
                  &            & \iconm-Ans         & \iconm-Full  \\
                  &            &  An $|$ Sp         & An+Sf $|$ Sp+Sf \\

        \midrule
        \multirow{2}{*}{\rotatebox[origin=c]{90}{\parbox[c]{1.7cm}{\centering \Mh Models}}}
                  & EE         & 40.7 $|$ 69.4      & 24.0 $|$ 25.6 \\
                  & SA         & 52.3 $|$ 75.2      & 34.8 $|$ 42.1 \\
                  & Ex(EE)     & 46.4 $|$ 78.1      & 32.2 $|$ 44.2 \\
                  & Ex(SA)     & 49.0 $|$ 80.6      & 32.2 $|$ 44.3 \\

        \midrule
        \multirow{2}{*}{\rotatebox[origin=c]{90}{\parbox[c]{1.4cm}{\centering Artifact Models}}}
                  & 1-Para     &   35.7 $|$ \mpt    & \p2.3 $|$ \mpt  \\
                  & C-only     &  \p3.7 $|$ \p0.0   & \p1.6 $|$ \p1.1 \\
                  & Q-only     &  \p4.6 $|$ \mpt    & \p0.0 $|$ \mpt \\

        \bottomrule
    \end{tabular}
    \caption{
        \mf is harder (top row) and less cheatable (bottom row) than \ma. Note: \mf has a stricter metric that operates over instance pairs (\S\ref{subsubsec:datasets}:metrics).
    }
    \label{table:musiq-full-better}
\end{table}

Intuitively, because the answerable and unanswerable instances are very similar but have different labels, it's difficult for models to do well on both instances if they learn to rely on shortcuts~\cite{kaushik2019learning,gardner2020evaluating}. All artifact-based models barely get any An+Sf or Sp+Sf score. For all \mh models too, the An drops by 14-17 pts and Sp by 33-44 pts.

\subsection{Dataset Construction Steps are Valuable}
\label{subsec:pipeline-valuable}

Next, we show that the key steps of our dataset construction pipeline (\S\ref{sec:construction-pipeline}) are valuable.

\paragraph{Disconnection Filter (step 3).}

To assess the effect of Disconnection Filter (DF), we ablate it from the pipeline, ie., skip the filtering composable 2-hop questions to connected 2-hop questions. As we don't have human-generated composed questions for the resulting questions, we use a seq2seq BART-large model that's trained (using \data) to compose questions from input decomposition DAG. For a fair comparison, we randomly subsample train set from ablated pipeline to be of the same size as the original train set.

\begin{table}[ht]
    \centering
    \small
    \begin{tabular}{lccc}\toprule
        & 1-Para & C-only & EE \\
        & An $|$ Sp & An $|$ Sp & An $|$ Sp \\
        \midrule
        \iconm                   & 32.0 $|$ \mpt & \p3.4 $|$ \p0.0 & 42.3 $|$ 67.6 \\
        \iconm $\setminus$ DF    & 59.2 $|$ \mpt & \p8.6 $|$  22.4 & 60.6 $|$ 71.1 \\
        \iconm $\setminus$ RL    & 85.1 $|$ \mpt &  69.5 $|$  42.3 & 87.3 $|$ 79.3 \\
        \bottomrule
    \end{tabular}
    \caption{
    Disconnection Filter (DF, step 5) and Reduced Train-Test Leakage (RL, step 3) of \data pipeline are crucial for its difficulty (EE model) and less cheatability (1-Para and C-only models).}
    \label{table:musiq-pipeline-ablations}
\end{table}

Table~\ref{table:musiq-pipeline-ablations} shows that DF is crucial for increasing difficulty and reducing cheatability of the dataset. Without DF, both \mh and artifact-based models do much better on the resulting datasets.

\paragraph{Reduced Train-Test Leakage (step 5).}

To assess the effect of Reduced train-test Leakage (RL), we create a dataset the traditional way, with a random partition into train, validation, and test splits. For uniformity, we ensure the distribution of 2-4 hop questions in development set of the resulting dataset from both ablated pipelines remains the same as in the original development set. Like DF ablation, we also normalize train set sizes.

Table~\ref{table:musiq-pipeline-ablations} shows that without a careful split, the dataset is highly solvable by \mh models (An=87.3). Importantly, most of this high score can also be achieved by artifact-based models: 1-para (An=85.1) and C-only (An=69.5), revealing the high cheatability of such a split.

\paragraph{Harder Distractors (step 7).}

To assess the effect of distractors in \ma, we create 4 variations. Two vary the number of distractors: (i) 10 paragraphs and (ii) 20 paragraphs; and two vary the source: (i) Full wikipedia (FW)\footnote{We used the Wikipedia corpus from~\citet{petroni2021kilt}.} and (ii) gold context paragraphs from the good \sh questions from step 1. We refer to the last setting as \textbf{positive distractors} (PD), as these paragraphs are likely to appear as supporting (positive) paragraphs in our final dataset. 

\begin{table}[ht]
    \centering
    \small
    \setlength{\tabcolsep}{4pt}
    \begin{tabular}{p{0.6cm}cccc}\toprule

        Ctxt & Corpus & 1-Para & C-Only & EE \\
        & &  An $|$ Sp & An $|$ Sp & An $|$ Sp \\

        \midrule
        \iconpm 10    &  FW     & 42.5 $|$ \mpt &  12.5 $|$ 77.7   & 57.2 $|$ 87.6  \\
        \iconpm 10    &  PD     & 28.0 $|$ \mpt & \p5.5 $|$ 34.6   & 54.1 $|$ 80.2  \\

        \midrule
        \iconpm 20    &  FW     & 41.7 $|$ \mpt &  12.4 $|$  66.4  & 50.3 $|$ 80.8  \\
        \iconm  20    &  PD     & 32.0 $|$ \mpt & \p3.4 $|$ \p0.0  & 42.3 $|$ 67.6  \\

        \bottomrule
    \end{tabular}
    \caption{Positive Distractors (PD) are more effective than using Full Wikipedia (FW) for choosing distractors, as shown by lower scores of models. The effect of using PD is more pronounced when combined with the use of 20 (rather than 10) distractor paragraphs.}
    \label{table:positive-distractors-better}
\end{table}

Table~\ref{table:positive-distractors-better} shows that all models find PD significantly harder than FW. In particular, PD makes support identification extremely difficult for C-only, whereas Table~\ref{table:musiq-vs-others} showed that C-only succeeds on \hq and \tw to a high degree (67.6 and 92.0 Sp). This would have also been true for \ma (66.4 Sp) had we used Wikipedia as the distractor construction corpus like \hq and \tw. This underscores the value of selecting the right corpus for distractor selection, and ensuring distributional shift can't be exploited to bypass reasoning.\footnote{Our \sh datasets are wikipedia-based, and we ensured retrieved contexts from FW are 20-300 words, like PD.} 

Second, using 20 paragraphs instead of 10 makes the dataset more difficult and less cheatable. Interestingly, the effect is stronger if we use PD, indicating the synergy between two approaches to create challenging distractors.

\subsection{Potential Avenues for Improvement}
\label{subsec:misc-improvements}

\paragraph{Better Decomposition.}

We train our EX(SA) model using ground-truth decompositions. On \ma, (An, Sp) improve by (9.4, 7.3) pts, and on \mf, (An+Sf, Sp+Sf) improve by (7.3, 6.9) pts. The improvements with the EX(EE) model are slightly lower. This shows that although improving question decomposition will be helpful, it's insufficient to reach human parity on the dataset.

\paragraph{Better Transformer.}

While Longformer can fit long context, there are arguably more effective pretrained transformers for shorter input, e.g., T5. Moreover, since T5 uses relative position embeddings, it can be used for longer text, although at a \href{https://github.com/huggingface/transformers/issues/5204#issuecomment-648045999}{significant memory and computation cost}. We managed to train SA with T5-large on \data,\footnote{SA worked best for 7 selected paragraphs, where the answerer (T5) had to process $\sim$1100 wordpieces on average.} but didn't use it for the rest of our experiments because of high computational cost. Over Longformer SA, T5 SA showed a modest improvement of (6.1, 0.7) on \ma and (1.7, 2.0) on \mf.

\section{Conclusion}
Constructing \mh datasets is a tricky process. It can introduce shortcuts and artifacts that models can exploit to circumvent the need for \mh reasoning. A bottom-up process of constructing \mh from \sh questions allows systematic exploration of a large space of \mh candidates and greater control over which questions we compose. We showed how to use such a carefully controlled process to create a challenging dataset that, by design, requires connected reasoning by reducing potential reasoning shortcuts, minimizing train-test leakage, and including harder distractor contexts. Empirical results show that \ma has a substantially higher human-model gap and is significantly less cheatable via disconnected reasoning than previous datasets. The dataset also comes with unanswerable questions, and question decompositions which we hope spurs further work in developing models that get right answers for the right reasons.

\subsection*{Acknowledgments}

The authors thank the action editor and reviewers for their valuable feedback.
This work was supported in part by the National Science Foundation under grant IIS-1815358.

\bibliography{MuSiQue}
\bibliographystyle{acl_natbib}

\newpage

\appendix
\section{Appendix}
\label{sec:appendix}

\begin{figure*}
    \centering
	\includegraphics[width=\textwidth]{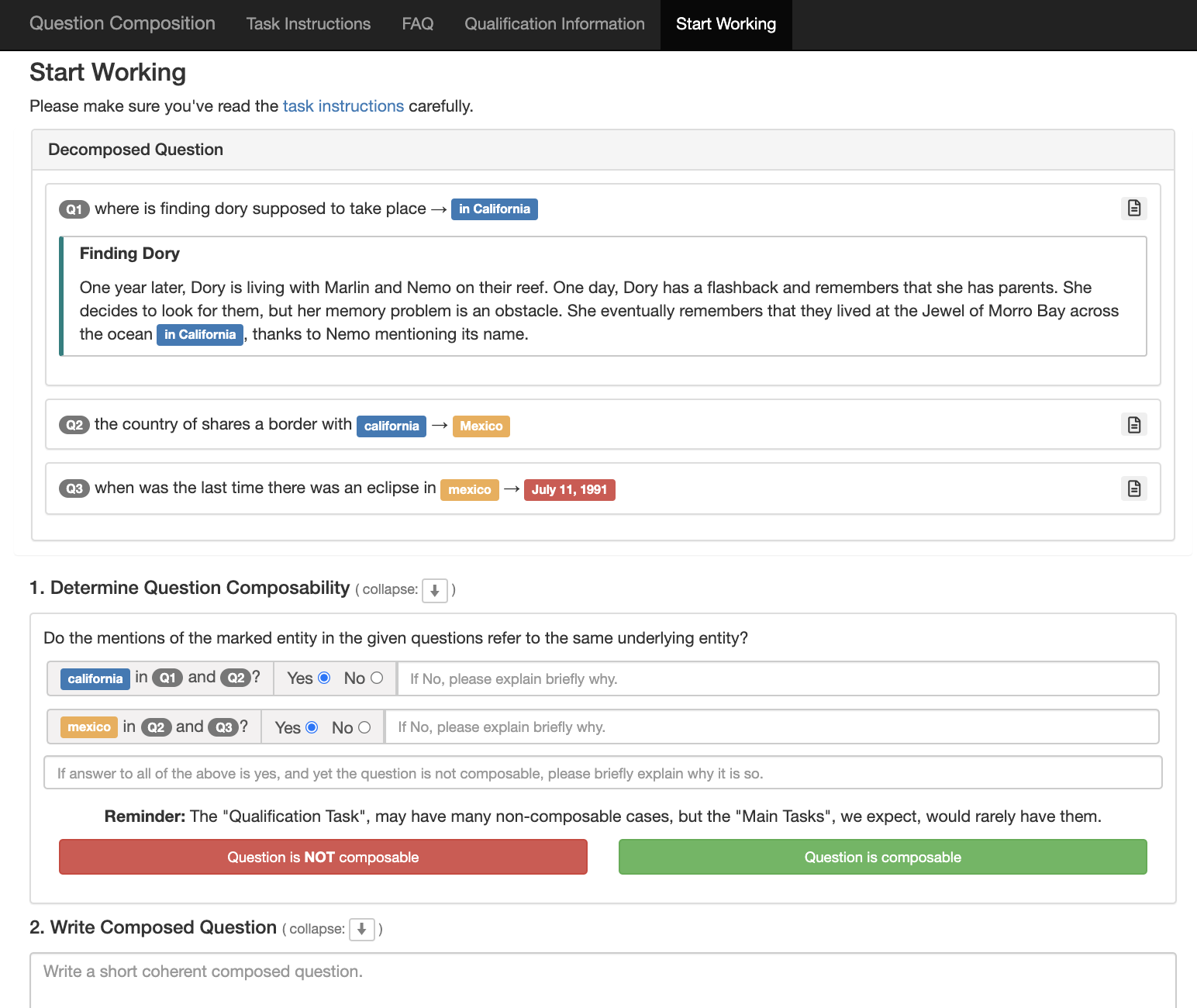}
	\caption{Annotation interface used for the question composition task. Workers could see decomposition graph and passage associated with subquestions.}
	\label{fig:composition-annotation-interface}
\end{figure*}

\begin{figure*}
    \centering
	\includegraphics[width=\textwidth]{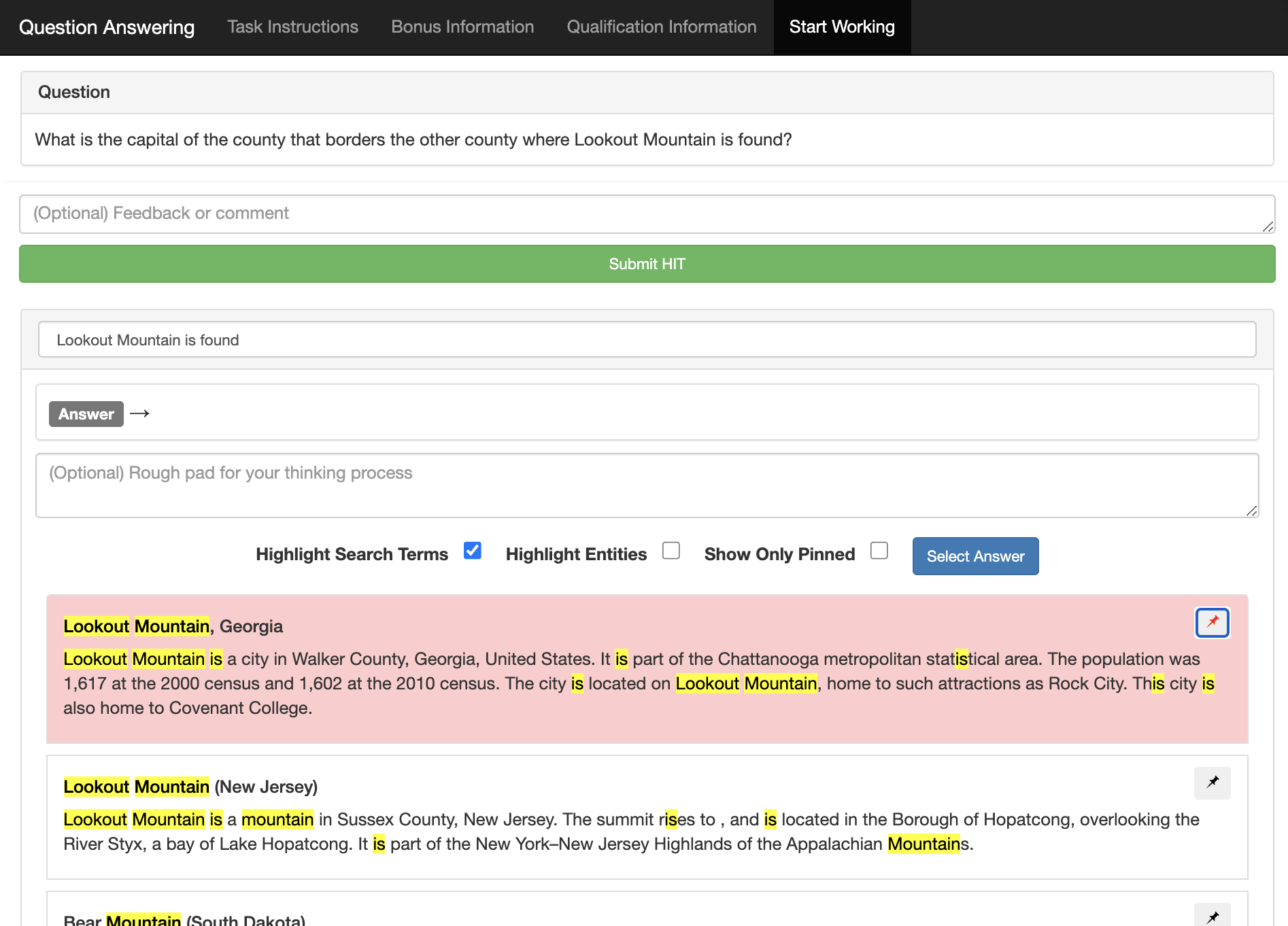}
	\caption{Annotation interface used for establishing human scores on \dataans, \hpqa and \twowikifull.}
	\label{fig:validation-annotation-interface}
\end{figure*}

Figure~\ref{fig:composition-annotation-interface} shows our annotation interface for the question composition task. Figure~\ref{fig:validation-annotation-interface} shows our annotation interface for establishing human scores on \dataans, \twowikifull and \hpqa.

\end{document}